\documentclass[10pt,twocolumn,letterpaper]{article}

\usepackage{cvpr}

\usepackage{graphicx,amsmath, amssymb, booktabs}
\usepackage[pagebackref,breaklinks,colorlinks]{hyperref}
\usepackage[capitalize]{cleveref}
\crefname{section}{Sec.}{Secs.}
\Crefname{section}{Section}{Sections}
\Crefname{table}{Table}{Tables}
\crefname{table}{Tab.}{Tabs.}

\usepackage{enumitem, graphicx, multirow, amsfonts, amsmath, verbatim, algorithm, algorithmicx, algpseudocode, booktabs, array, arydshln, times, epsfig, amssymb, pifont, capt-of, wrapfig, lipsum, makecell, tabulary, etoolbox, subcaption, xcolor}
\newcommand{\tablestyle}[2]{\setlength{\tabcolsep}{#1}\renewcommand{\arraystretch}{#2}\centering\footnotesize}

\def\cvprPaperID{1961}
\def\confName{CVPR}
\def\confYear{2023}

\begin{document}
\title{Tell Me What Happened: Unifying Text-guided Video Completion\\via Multimodal Masked Video Generation}
\author{Tsu-Jui Fu$^\text{1}$, Licheng Yu$^\text{2}$, Ning Zhang$^\text{2}$, Cheng-Yang Fu$^\text{2}$,\\Jong-Chyi Su$^\text{3}$, William Yang Wang$^\text{1}$, Sean Bell$^\text{2}$\\$^\text{1}$UC Santa Barbara~~$^\text{2}$Meta~~$^\text{3}$NEC Laboratories America\\
{\tt \small \{tsu-juifu, william\}@cs.ucsb.edu~~jcsu@nec-labs.com}\\
{\tt \small \{lichengyu, ningzhang, chengyangfu, seanbell\}@meta.com}\\
}
\maketitle

\begin{abstract}
Generating a video given the first several static frames is challenging as it anticipates reasonable future frames with temporal coherence.
Besides video prediction, the ability to rewind from the last frame or infilling between the head and tail is also crucial, but they have rarely been explored for video completion.
Since there could be different outcomes from the hints of just a few frames, a system that can follow natural language to perform video completion may significantly improve controllability.
Inspired by this, we introduce a novel task, text-guided video completion (\texttt{TVC}), which requests the model to generate a video from partial frames guided by an instruction.
We then propose Multimodal Masked Video Generation (MMVG) to address this \texttt{TVC} task.
During training, MMVG discretizes the video frames into visual tokens and masks most of them to perform video completion from any time point.
At inference time, a single MMVG model can address all 3 cases of \texttt{TVC}, including video prediction, rewind, and infilling, by applying corresponding masking conditions.
We evaluate MMVG in various video scenarios, including egocentric, animation, and gaming.
Extensive experimental results indicate that MMVG is effective in generating high-quality visual appearances with text guidance for \texttt{TVC}.
\end{abstract}

\vspace{-2ex}
\section{Introduction}
\begin{figure}[t]
\centering
    \includegraphics[width=\linewidth]{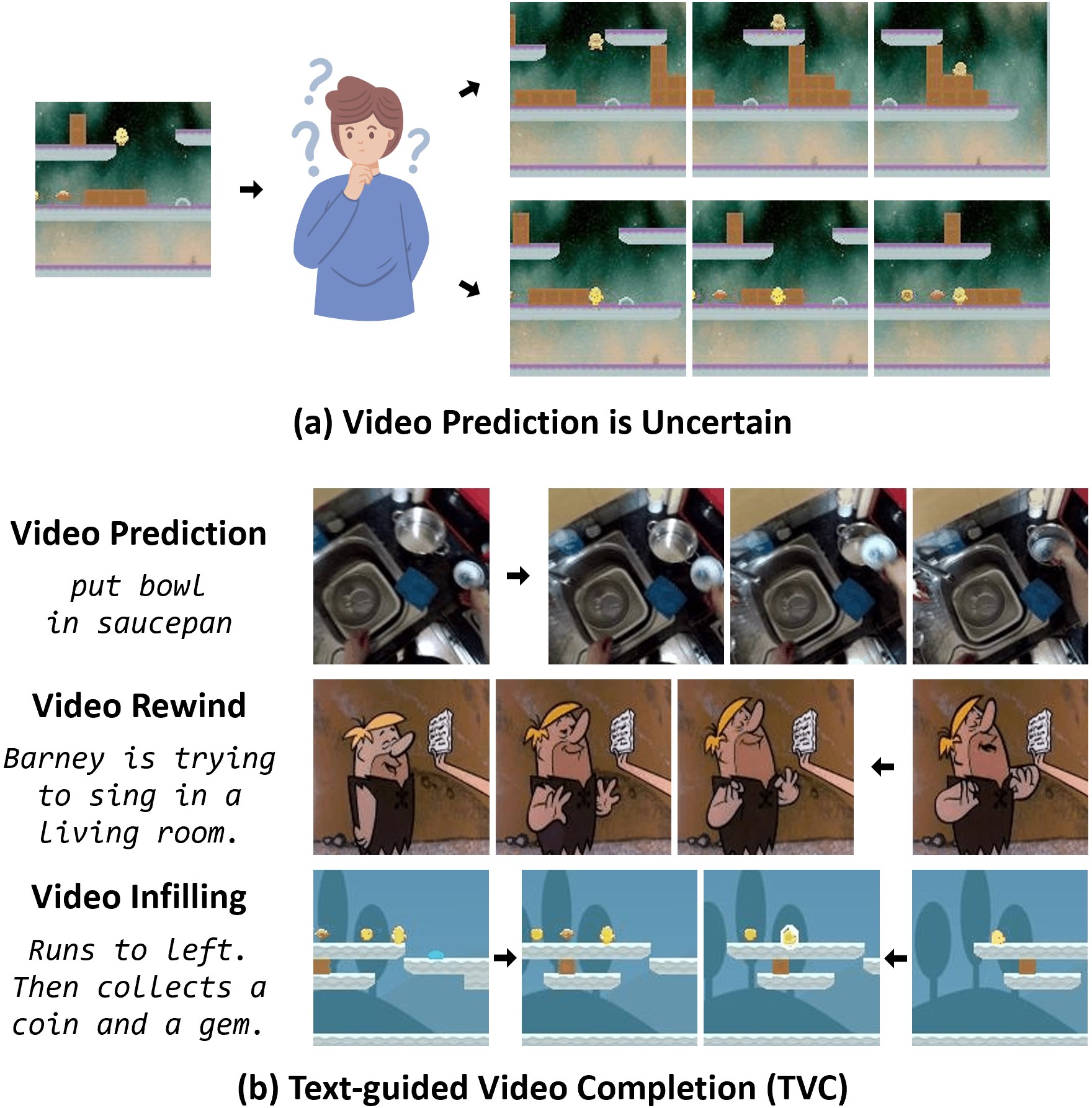}
    \vspace{-4ex}
    \caption{The introduced text-guided video completion (\texttt{TVC}) task. 
    (a) Video prediction may have different outcomes without text guidance.  
    (b) \texttt{TVC} performs video completion from the first frame (\textbf{prediction}), the last frame (\textbf{rewind}), or both (\textbf{infilling}), guided by the textual description.}
    \vspace{-3ex}
    \label{fig:tvc}
\end{figure}

Generative video modeling~\cite{denton2018video-generation,srivastava2015video-lstm,weissenborn2020autogressive-video} has made great progress, which first succeeds in unconditional video generation~\cite{kalchbrenner2017video-pixel,saito2018tgan}. 
More recently, video prediction~\cite{hao2017controllable-video-generation,hsieh2018video-prediction,mathieu2018video-prediction} has been trying the controllable setting, which anticipates the future by completing a video from the past frames or a static starting image~\cite{hu2018frame-to-video,zhang2020dtv-net}. 
However, video prediction may produce various outcomes, which makes it difficult to meet human expectations. 
For the example in Fig.~\ref{fig:tvc}(a), the game agent can keep jumping to the right or move back and turn left. The limited guidance from only the first frame is insufficient to tell the intention. For humans, language is the most straightforward way of communication.
If a system can follow an instruction to accomplish video completion, it will significantly improve its controllability and make a vast application impact. 
On the other hand, compared with video prediction, video rewind and infilling have been rarely studied~\cite{hoppe2022ramvid,voleti2022mcvd}, but they are also crucial.
Breaking the limitation of chronological guidance should make the visual guidance more flexible, leading to a general video completion.

We thus introduce a novel task, text-guided video completion (\texttt{TVC}), where the partial frames and a given instruction jointly guide the video generation. 
As illustrated in Fig.~\ref{fig:tvc}(b), we consider three scenarios of video completion: \textbf{prediction} from the first frame, \textbf{rewind} from the last frame, and \textbf{infilling} between the head and tail. 
The missing (to-be-completed) event should follow the textual instruction.  
Compared to generating content from scratch~\cite{li2018t2v,wu2022nuwa}, \texttt{TVC} requests models to understand the given visual and textual guidance before generation, which better mimics how human imagines after seeing and listening in our daily lives.

\begin{figure*}[t]
\centering
    \includegraphics[width=\linewidth]{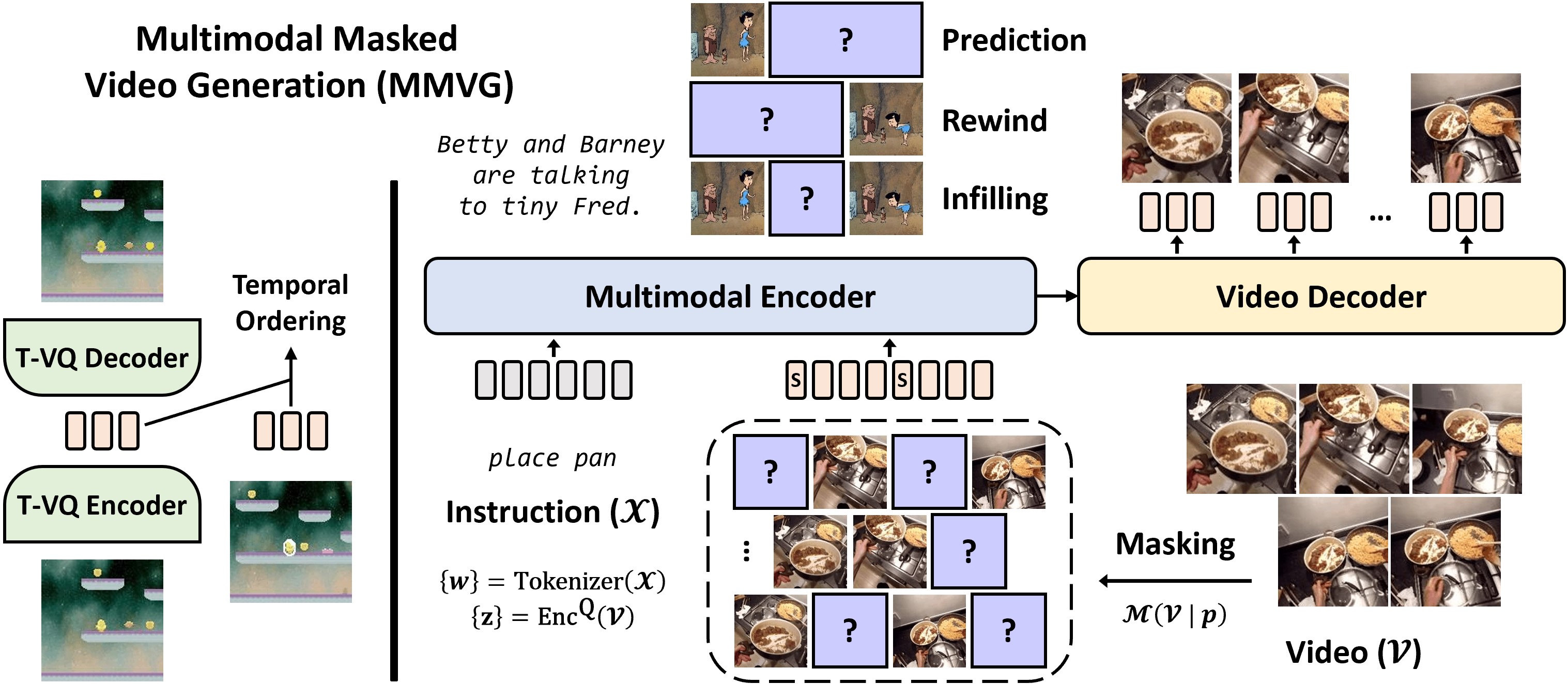}
    \vspace{-4ex}
    \caption{An overview of our Multimodal Masked Video Generation (MMVG). We present temporal-aware VQGAN (T-VQ) for discrete visual representation. MMVG considers the instruction $\mathcal{X}$ and partial frames of the video $\mathcal{V}$ from diverse time points through masking, learning to generate the complete $\mathcal{V}$. In this way, a single trained MMVG can perform all prediction, rewind, and infilling tasks.}
    \vspace{-3ex}
    \label{fig:mmvg}
\end{figure*}

To tackle \texttt{TVC}, we present Multimodal Masked Video Generation (MMVG) to perform video completion. 
Specifically, we represent the video frames as discrete visual tokens by temporal-aware VQGAN~\cite{oord2017vq-vae,esser2021taming}. 
One key challenge is to deal with the video frames that are not presented in chronological (\textit{e.g.}, the last frame for rewind).
Different from autoregressive models~\cite{ge2022tats,wu2021godiva} that only condition on the previous frames, MMVG carries out video completion in an encoder-decoder manner. 
Specifically, we propose a masking strategy that masks different parts of the video and feeds them as the input to the multimodal encoder with the instruction. 
As shown in Fig.~\ref{fig:mmvg}, we allow MMVG to consider the visual hints from different time points, and the decoder learns to produce the full target video. 
By varying the masking conditions (including the cases of only the first or last frame being accessible), a single MMVG can address all \texttt{TVC} tasks, including video prediction, rewind, and infilling. 
Moreover, learning the recovery from partial frames also empowers MMVG with a strong temporal coherence, contributing to better generative video quality.

We consider videos in diverse scenarios for the \texttt{TVC} evaluation. 
There are Kitchen~\cite{damen2018kitchen}, Flintstones~\cite{gupta2018flintstones}, and MUGEN~\cite{hayes2022mugen} corresponding to the egocentric, animation, and gaming scenes.
The model should generate videos such as performing kitchen activities in the first-person view, making characters act the assigned behavior, or imitating an agent playing game. 
All should be guided with the first/last (or both) frame(s) and controlled through the given human instructions.
We also compare MMVG with previous methods~\cite{yu2022di-gan,ge2022tats,rakhimov2020lvt,yan2021video-gpt} on UCF-101~\cite{soomro2012ucf101} and BAIR~\cite{ebert2017bair} for the classic video generation/prediction tasks.

Experimental results demonstrate that instruction is necessary to make video completion controllable, MMVG can address all three \texttt{TVC} tasks,  and our proposed masking strategy enhances the temporal modeling, which further benefits general video generation/prediction.
In summary, our contributions are three-fold:
\begin{itemize}[noitemsep, leftmargin=*, topsep=2pt]
    \item We introduce \texttt{TVC} to generate a video from partial frames and control the temporal dynamics via natural language, where our video completion includes 3 cases: prediction, rewind, and infilling.
    \item We propose MMVG with an effecitve masking strategy to address all \texttt{TVC} tasks through a single training.
    \item Extensive experiments show that our MMVG can handle various types of video completion as well as video generation/prediction. We believe \texttt{TVC} can become a new topic in vision-and-language research.
\end{itemize}

\section{Related Work}
\noindent \textbf{Video Generation/Prediction.~}
Video generation aims to synthesize diverse videos from latent inputs~\cite{acharya2018video-gan,tulyakov2018moco-gan,vondrick2016video-gen}. Various generative modelings have shown promising results, including generative adversarial networks (GAN)~\cite{goodfellow2015gan,tian2021moco-gan-hd,clark2019dvd-gan,yu2022di-gan}, autoregressive transformers~\cite{vaswani2017transformer,yan2021video-gpt,ge2022tats}, and denoising diffusion models~\cite{ho2020diffusion,dhariwal2021diffusion,ho2022video-diffusion}. Upon that, video prediction~\cite{babaeizadeh2018s2vp,rakhimov2020lvt,weissenborn2020vt,babaeizadeh2021fit-vid,gupta2022mask-vit}, which considers past frames to anticipate future observations, should maintain temporal dynamics from static images. Though the overall idea is also to complete a video from partial frames, other tasks, such as rewind and infilling~\cite{xu2020video-infilling,voleti2022mcvd,hoppe2022ramvid}, are not extensively explored. In this paper, we introduce \texttt{TVC} to comprehensively investigate the ability of video completion and make it more maneuverable via textual description.

\vspace{0.5ex}
\noindent \textbf{Text-to-Image/Video Genreation.~}
Generating visual content from language~\cite{nguyen2017plug-play,cheng2013image-spirit,tan2019text2scene} has a vast application value in creative visual design. Previous works rely on the GANs framework~\cite{mirza2014cgan} to produce images~\cite{reed2016t2i,xu2108attn-gan,qiao2019mirror-gan,el-nouby2019geneva,fu2020sscr,fu2022ldast} or videos~\cite{marwah2017att-video-gen,pan2017create-tell,li2018t2v}, conditioned on text. With large-scale datasets~\cite{sharma2108cc,schuhmann2021laion,wang2019vatex,bain2021frozen}, recent pre-trained models can generate high-quality natural images from open-domain description through discrete visual tokens~\cite{oord2017vq-vae,esser2021taming,ramesh2021dalle,ding2021cog-view,crowson2022vq-gan-clip,yu2022parti} or the diffusion process~\cite{ramesh2022dalle2,rombach2022latent-diffusion,nichol2022glide,saharia2022imagen}. Leveraging such techniques further extends to generate vivid videos~\cite{wu2021godiva,wu2022nuwa,hong2022cog-video,villegas2022phenaki,singer2022make-a-video,ho2022imagen-video}. However, those methods that depend on autoregressive generation can only be guided chronologically~\cite{hu2022make-it-move,han2022show-tell}. Besides, video diffusion models require a deterministic video length, which cannot consider diverse temporal durations. In contrast, MMVG can perform video completion in different lengths from arbitrary time points and address all \texttt{TVC} tasks just with a single training.

\vspace{0.5ex}
\noindent \textbf{Text-guided Video-to-Video.~}
Video inpainting~\cite{chang2019video-inpaint,kim2019video-inpaint,xu2019flow-guided}, segmentation reconstruction~\cite{wang2018v2v,wang2019few-shot-v2v}, or video style transfer~\cite{chen2017video-st,deng2021video-st,xia2021video-st} can be seen as a particular case of video-to-video synthesis (V2V). Even if text-guided V2V~\cite{fu2022lbve,bar-tal2022text2live,xu2022video-edit} can be controlled by language, it is still conditioned on a full video, where the temporal dynamics are usually provided. Different from that, \texttt{TVC} requires to regain the missing event from just partial guidance. It is more challenging since the model has to capture what happened from the instruction, maintain the temporal coherence among limited frames, and produce a complete video.

\section{Text-guided Video Completion (\texttt{TVC})}
\subsection{Task Definition}
We study the text-guided video completion (\texttt{TVC}) task to perform video completion from the first frame (prediction), the last frame (rewind), or the head and tail (infilling), conditioned on the textual instruction. 
During training, we have pairs of videos $\mathcal{V}$ and corresponding instructions $\mathcal{X}$. 
Specifically, $\mathcal{V}$ consists of $N$ frames as $\{v_1, v_2, ..., v_N\}$. 
Our goal is to train a unified model that generates the complete $\mathcal{V}$ given the partial frames from arbitrary time points and $\mathcal{X}$.

\subsection{Multimodal Masked Video Generation}
\noindent \textbf{Overview.~}
An overview of our Multimodal Masked Video Generation (MMVG) is illustrated in Fig.~\ref{fig:mmvg}. 
To model the video along with language, we propose temporal-aware VQGAN to represent a frame as visual tokens, which converts it into the same discrete space as the words. 
We present an effective masking strategy that masks different video parts for video completion learning. 
Those missing fragments are replaced by the unique [\texttt{SPAN}] tokens, and we consider the visual guidance from diverse time points. 
The multimodal encoder consumes the text and the partial-missing video, and the decoder learns to produce the complete video from arbitrary guided frames. 
By varying the masking conditions, MMVG learns to utilize the [\texttt{SPAN}] token and unifies all \texttt{TVC} tasks during the training.

\vspace{0.5ex}
\noindent \textbf{Temporal-aware Discrete Visual Tokens.~}
VQ-VAE~\cite{oord2017vq-vae} has shown promising capability in representing data as discrete tokens. 
VQGAN~\cite{esser2021taming} further models the prior distribution of the latent space via a transformer with the GAN training. 
If VQGAN is directly applied onto videos, it will ignore the inner temporal coherence and treat each frame as an independent image, resulting in an unsmooth video reconstruction.
Though TATS~\cite{ge2022tats} attempts to handle this by making $k$ consecutive frames altogether during VQ, it has to pre-define the constant $k$ before training. 
Such constraint forbids it from representing a frame at any timestamp.

To address it with flexibility, we propose temporal-aware VQGAN (T-VQ) to inject the temporal relationship into the latent representation. 
We first follow VQGAN to learn the target visual tokens $\text{z}_i$ by reconstructing a video frame $v_i$:
\begin{align} \label{eq:vq}
    \text{z}_i &= \text{q}(\text{Enc}^\text{Q}(v_i)~|~C), \notag \\
    \hat{v_i} &= \text{Dec}^\text{Q}(\text{z}_i), \notag \\
    \mathcal{L}_\text{VQ} &= \underbrace{||\hat{v_i}-v_i||_1}_{\text{reconstrcution}}+\underbrace{||\text{sg}[\text{Enc}^\text{Q}(v_i)]-C_{\text{z}_i}||_2^2}_{\text{codebook}} \\
    +& \underbrace{\beta||\text{sg}[C_{\text{z}_i}]-\text{Enc}^\text{Q}(v_i)||_2^2}_{\text{commit}} + \underbrace{||\mathcal{F}(\hat{v_i})-\mathcal{F}(v_i)||_1}_{\text{matching}}, \notag
\end{align}
where $\text{Enc}^\text{Q}$ and $\text{Dec}^\text{Q}$ are the VQ encoder and decoder. 
The discrete latent code $\text{z}_i$ is acquired from the quantization operation $\text{q}$~\cite{esser2021taming}, which adopts nearest neighbor search by the trainable codebook $C$.
We apply the straight-through estimator over the stop-gradient operation sg and use $\beta$ as $0.25$~\cite{oord2017vq-vae}.
We also append VGG~\cite{simonyan2015vgg} features matching $\mathcal{F}$ to stabilize the VQ loss $\mathcal{L}_\text{VQ}$~\cite{ge2022tats}. 
The adversarial training between the frame quality loss $\mathcal{L}_G$ and discrimination loss $\mathcal{L}_D$ are further calculated from the discriminator $D$:
\begin{equation}
\begin{split} \label{eq:gan}
    \mathcal{L}_G &= \log(1-D(\hat{v_i})),\\
    \mathcal{L}_D &= \log(1-D(\hat{v_i}))+\log(D(v_i)).
\end{split}
\end{equation}

To inject the temporal relationship into $\text{z}$, T-VQ is trained with the introduced contrastive temporal reasoning:
\begin{equation}
\begin{split} \label{eq:temporal}
    o_i &= \text{FC}^\text{T}(\text{z}_i, \text{z}_j), \\
    \mathcal{L}_\text{T} &= \text{BCELoss}(o_i, 0~\text{if $i>j$ else $1$}),
\end{split}
\end{equation}
where $j$ is a random frame from the same video. $\text{FC}^\text{T}$ is the MLP classifier, and BCELoss is the binary cross-entropy for before/after. 
Learning the temporal order from $\mathcal{L}_\text{T}$, $\text{z}$ facilitates an implicit temporal coherence, leading to smooth video modeling. 
Moreover, since $\text{z}$ represents a single image, it is flexible to support frames at arbitrary timestamps.

\vspace{0.5ex}
\noindent \textbf{Generation from Masked Video.~}
We propose the masking strategy $\mathcal{M}$ to obtain the masked videos $\overline{\mathcal{V}}$ from diverse time points. 
$\mathcal{M}$ masks out most video frames with the probability $p$ and replaces each fragment as a unique [\texttt{SPAN}] token. 
For example, $\mathcal{M}$ reserves the third and the fifth frame, and masks all the others over a video length of $5$:
\begin{equation}
    \overline{\mathcal{V}}: \{[\text{\texttt{S}}], v_3, [\texttt{S}], v_5\} = \mathcal{M}(\mathcal{V}~|~p).
\end{equation}
Our goal is to recover the missing part from $\overline{\mathcal{V}}$ and perform video completion, guided by the instruction $\mathcal{X}$. To model between the vision and language modalities, we apply our $\text{Enc}^\text{Q}$ over $\overline{\mathcal{V}}$ for the discrete visual tokens $\{[\text{\texttt{S}}], \text{z}_3, [\texttt{S}], \text{z}_5\}$. 
We also tokenize the text $\mathcal{X}$ into word tokens $\{w_i\}_{i=1}^L$ with the CLIP tokenizer~\cite{radford2021clip}, where $L$ is the length of $\mathcal{X}$. 
As in the same discrete space, MMVG can achieve cross-modal fusion by the multimodal encoder ($\text{Enc}^\text{M}$) through the self-attention mechanism as the transformer~\cite{vaswani2017transformer}:
\begin{equation}
\begin{split}
    f^w_i, f^v_j &= \text{LP}^w(w_i),\text{LP}^v(\text{z}_j) \\
    \{h\} &= \text{Enc}^\text{M}([\{f^w\}, \{f^v\}]),
\end{split}    
\end{equation}
where it obtains the features $f$ by the linear projection (LP), and $h$ is the hidden encoding features. 
We can also regard LP as the video/language embedder, which extracts the preliminary visual/linguistic features.

After encoding the language hint and the partial-missing video from $\text{Enc}^\text{M}$, our video decoder ($\text{Dec}^\text{M}$) learns to produce all frames for comprising the complete video. $\text{Dec}^\text{M}$ follows the vanilla autoregressive decoder, which first conducts self-attention over the past generated tokens and then predicts the discrete visual tokens as the video frame, conditioned on the encoded features $h$:
\begin{equation}
\begin{split} \label{eq:dec}
    \hat{\text{z}_t} &= \text{Dec}^\text{M}(\{\hat{\text{z}_1}, ..., \hat{\text{z}_{t-1}}\}~|~\{h\}), \\
    \mathcal{L}_t &= \text{CELoss}({\hat{\text{z}_t}, \text{z}_t}), \\
    \mathcal{L}_\text{M} &= \sum_{t=1}^{N} \mathcal{L}_t,
\end{split}
\end{equation}
where $\text{z}_t$ is the ground-truth tokens of the frame $v_t$ in the original $\mathcal{V}$. We calculate the video decoding loss $\mathcal{L}_\text{M}$ by the cross-entropy (CELoss) to learn video generation as classification. 
Our $\text{Dec}^\text{M}$ is built upon VideoSwin~\cite{liu2022video-swin}, which has shown a strong visual perception on various video understanding tasks. 
The 3D-shifted windows~\cite{liu2021swin} consider different levels of spatial-temporal attention, and each window models blocks of video patches across $T'$ consecutive frames. 
To ensure the same dimension for video generation in $\text{Dec}^\text{M}$, we remove the temporal down-sampling layer. In the end, we can utilize $\text{Dec}^\text{Q}$ to reconstruct all the frames as our completed videos $\hat{\mathcal{V}}$:
\begin{equation}
    \hat{\mathcal{V}} = \text{Dec}^\text{Q}(\{\hat{\text{z}}\}_{t=1}^{N}).
\end{equation}

\begin{algorithm}[t]
\small
    \begin{algorithmic}[1]
        \While{TRAIN\_T-VQ}
            \State $\mathcal{V}$ $\gets$ sample video
            \State $\text{z}_i$ = $\text{q}(\text{Enc}^\text{Q}(v_i)~|~C)$
            \State $\hat{v_i}$ = $\text{Dec}^\text{Q}(\text{z}_i)$
            \State $o_i$ = $\text{FC}^\text{T}(\text{z}_i, \text{z}_j)$
            \Comment{randomly sampled frame $j$}
            \State $\mathcal{L}_\text{VQ}$, $\mathcal{L}_G$ $\gets$ reconstruction, frame quality loss
            \Comment{Eq.~\ref{eq:vq}\&\ref{eq:gan}}
            \State $\mathcal{L}_\text{T}$ $\gets$ temporal ordering loss \Comment{Eq.~\ref{eq:temporal}}
            \State Update T-VQ by minimizing $\mathcal{L}_\text{VQ}$+$\mathcal{L}_G$+$\mathcal{L}_\text{T}$
            \State $\mathcal{L}_D$ $\gets$ discrimination loss \Comment{Eq.~\ref{eq:gan}}
            \State Update D by maximizing $\mathcal{L}_D$
        \EndWhile
        \\
        \While{TRAIN\_MMVG}
            \State $\mathcal{V}$, $\mathcal{X}$, $p$ $\gets$ sample video/instruction/probability
            \State $\overline{\mathcal{V}}$:~$\{v_a, \text{[\texttt{S}]}, v_b, ... \}$ = $\mathcal{M}(\mathcal{V}~|~p)$
            \Comment{diverse guided frames}
            \State $\{\text{z}_a, [\texttt{S}], \text{z}_b, ...\}, \{w\}$ = $\text{Enc}^\text{Q}$($\overline{\mathcal{V}}$), Tokenizer($\mathcal{X}$)
            \State $\{h\}$ = $\text{Enc}^\text{M}([\{w\}, \{\text{z}_a, \text{[\texttt{S}]}, \text{z}_b, ... \}])$
            \For{$t \gets 1$ to $N$}
                \State $\hat{\text{z}_t}$ = $\text{Dec}^\text{M}(\{\text{z}_1, ..., \text{z}_{t-1}\}~|~\{h\})$
                \Comment{teacher-forcing}
                \State $\mathcal{L}_t$ $\gets$ video decoding loss
                \Comment{Eq.~\ref{eq:dec}}
            \EndFor
            \State $\hat{\mathcal{V}}$ = $\text{Dec}^\text{Q}(\{\hat{\text{z}}_{t=1}^N\})$
            \State $\mathcal{L}_\text{M}$ = $\sum^{N}_{t=1} \mathcal{L}_t$
            \State Update MMVG by minimizing $\mathcal{L}_\text{M}$
            \State $p$ $\gets$ update masking probability
            \Comment{Eq.~\ref{eq:p}}
        \EndWhile
    \end{algorithmic}
    \caption{Mutlimodal Masked Video Generation}
    \label{algo:learning}
\end{algorithm}

\begin{table*}[t]
\centering \tablestyle{2pt}{1.1}
\resizebox{\linewidth}{!}{
    \begin{tabular}{ccccccccccccccccccccccccccccc}
        \toprule
        ~ & ~ & ~ & \multicolumn{8}{c}{\textbf{\texttt{TVP}rediction}} & ~ & \multicolumn{8}{c}{\textbf{\texttt{TVR}ewind}} & ~ & \multicolumn{8}{c}{\textbf{\texttt{TVI}nfilling}} \\
        \cmidrule{4-11} \cmidrule{13-20} \cmidrule{22-29}
        ~ & ~ & ~ & \multicolumn{2}{c}{\textbf{Kitchen}} & ~ & \multicolumn{2}{c}{\textbf{Flintstones}} & ~ & \multicolumn{2}{c}{\textbf{MUGEN}} & ~ & \multicolumn{2}{c}{\textbf{Kitchen}} & ~ & \multicolumn{2}{c}{\textbf{Flintstones}} & ~ & \multicolumn{2}{c}{\textbf{MUGEN}} & ~ & \multicolumn{2}{c}{\textbf{Kitchen}} & ~ & \multicolumn{2}{c}{\textbf{Flintstones}} & ~ & \multicolumn{2}{c}{\textbf{MUGEN}} \\
        \cmidrule{4-5} \cmidrule{7-8} \cmidrule{10-11} \cmidrule{13-14} \cmidrule{16-17} \cmidrule{19-20} \cmidrule{22-23} \cmidrule{25-26} \cmidrule{28-29}
        Method & Text & ~ & FVD$\downarrow$ & RCS$\uparrow$ & ~ & FVD$\downarrow$ & RCS$\uparrow$ & ~ & FVD$\downarrow$ & RCS$\uparrow$ & ~ & FVD$\downarrow$ & RCS$\uparrow$ & ~ & FVD$\downarrow$ & RCS$\uparrow$ & ~ & FVD$\downarrow$ & RCS$\uparrow$ & ~ & FVD$\downarrow$ & RCS$\uparrow$ & ~ & FVD$\downarrow$ & RCS$\uparrow$ & ~ & FVD$\downarrow$ & RCS$\uparrow$ \\
        \midrule
        FILM~\cite{reda2022film} & \ding{55} & ~ & - & - & ~ & - & - & ~ & - & - & ~ & - & - & ~ & - & - & ~ & - & - & ~ & 250.2 & 56.1 & ~ & 352.7 & 51.4 & ~ & 538.8 & 5.9 \\
        VideoMAE~\cite{tong2022video-mae} & \ding{55} & ~ & 328.9 & 47.6 & ~ & 317.5 & 55.6 & ~ & 548.7 & 7.0 & ~ & 365.9 & 48.2 & ~ & 335.5 & 55.9 & ~ & 545.2 & 7.1 & ~ & 246.9 & 54.7 & ~ & 211.5 & 60.6 & ~ & 494.9 & 7.8 \\
        TATS~\cite{ge2022tats} & \ding{55} & ~ & 106.9 & \underline{64.4} & ~ & 127.5 & 60.3 & ~ & 376.5 & 7.1 & ~ & \underline{107.7} & \underline{62.7} & ~ & 127.6 & \underline{60.2} & ~ & \underline{350.8} & \underline{7.2} & ~ & 71.5 & 72.7 & ~ & \underline{119.5} & \underline{66.7} & ~ & \underline{328.2} & \underline{8.4} \\
        MMVG$^\text{U}$ & \ding{55} & ~ & \underline{105.6} & 63.3 & ~ & \underline{124.8} & \underline{60.5} & ~ & \underline{374.5} & \underline{7.2} & ~ & 109.8 & 62.6 & ~ & \underline{124.3} & 59.7 & ~ & 356.4 & 7.0 & ~ & \underline{71.5} & \underline{73.4} & ~ & 121.8 & 66.3 & ~ & 328.4 & 7.8 \\
        MMVG$^\text{S}$ & \ding{55} & ~ & \textbf{103.8} & \textbf{64.5} & ~ & \textbf{123.8} & \textbf{60.8} & ~ & \textbf{369.4} & \textbf{7.3} & ~ & \textbf{105.9} & \textbf{63.6} & ~ & \textbf{123.8} & \textbf{60.5} & ~ & \textbf{347.8} & \textbf{7.2} & ~ & \textbf{68.5} & \textbf{73.6} & ~ & \textbf{118.5} & \textbf{67.9} & ~ & \textbf{324.3} & \textbf{8.4} \\
        \midrule
        TATS~\cite{ge2022tats} & \ding{51} & ~ & 87.2 & 66.3 & ~ & 115.9 & 70.6 & ~ & 90.1 & 67.9 & ~ & 89.8 & 63.3 & ~ & 116.3 & 70.4 & ~ & \underline{89.8} & \underline{68.7} & ~ & \underline{57.4} & 77.6 & ~ & 95.8 & 78.2 & ~ & \underline{58.9} & \underline{73.6} \\
        MMVG$^\text{U}$ & \ding{51} & ~ & \underline{80.2} & \underline{68.4} & ~ & \underline{108.2} & \underline{72.9} & ~ & \underline{84.8} & \underline{70.2} & ~ & \underline{83.2} & \underline{66.9} & ~ & \underline{113.2} & \underline{71.6} & ~ & 93.1 & 68.4 & ~ & 59.8 & \underline{77.8} & ~ & \underline{92.8} & \underline{78.3} & ~ & 59.2 & 73.2 \\
        MMVG$^\text{S}$ & \ding{51} & ~ & \textbf{75.6} & \textbf{68.8} & ~ & \textbf{106.3} & \textbf{73.7} & ~ & \textbf{83.3} & \textbf{71.1} & ~ & \textbf{79.7} & \textbf{68.1} & ~ & \textbf{107.2} & \textbf{72.9} & ~ & \textbf{88.7} & \textbf{70.0} & ~ & \textbf{56.0} & \textbf{78.1} & ~ & \textbf{91.6} & \textbf{79.6} & ~ & \textbf{57.2} & \textbf{74.1}  \\
        \bottomrule
    \end{tabular}
}
    \vspace{-2ex}
    \caption{Results of \texttt{TVC}, including prediction, rewind, and infilling, on Kitchen~\cite{damen2018kitchen}, Flintstones~\cite{gupta2018flintstones}, and MUGEN~\cite{hayes2022mugen}. TATS~\cite{ge2022tats} requires specific training to support different tasks. We further train the unified MMVG$^\text{U}$ for each specific task as MMVG$^\text{S}$.}
    \label{table:main}
    \vspace{-3ex}
\end{table*}

By varying the masking conditions through $\mathcal{M}$, MMVG learns how to complete a video from partial frames $\overline{\mathcal{V}}$ at arbitrary time points with the text, which overcomes the limitation of chronological guidance. 
To make $\mathcal{M}$ more effective, we apply an adaptive probability $p$ instead of random sampling every time. Each video $\mathcal{V}$ keeps its own $p$, and all frames are equally initialized in the beginning. 
Based on the prediction error, we adjust the masking probability $p_t$ of the $t$-th frame:
\begin{equation} \label{eq:p}
    p_t = p_t + \alpha ((\frac{\mathcal{L}_t}{\mathcal{L}_M} \sum p) - p_t),
\end{equation}
where $\alpha$ is the adjusting rate. 
A larger video decoding loss $\mathcal{L}_t$ indicates that the $t$-th frame is more difficult to recover. MMVG learns more from those challenging cases and can bring better generative quality for video completion.

\vspace{0.5ex}
\noindent \textbf{Unifying \texttt{TVC} during Inference.~}
After training with text and partial-missing video, MMVG learns to perform video completion over [\texttt{SPAN}] tokens. 
Then for inference, $\text{Enc}^\text{M}$ takes the following as its input to support different tasks:
\begin{itemize}[noitemsep, leftmargin=*, topsep=0pt]
    \item \texttt{TVP}rediction: [$\{w\}, \{\text{z}_1, \text{[\texttt{SPAN}]}$\}]
    \item \texttt{TVR}ewind: [$\{w\}, \{\text{[\texttt{SPAN}]}, \text{z}_N$\}]
    \item \texttt{TVI}nfilling: [$\{w\}, \{\text{z}_1, \text{[\texttt{SPAN}]}, \text{z}_N$\}]
\end{itemize}
In this way, a single trained MMVG can unify all \texttt{TVC} tasks without the specific downstream fine-tuning.

\subsection{Learning of MMVG}
Algo.~\ref{algo:learning} illustrates the learning process of our proposed MMVG for \texttt{TVC}. We first train T-VQ over video frames for discrete visual tokens with contrastive temporal reasoning. 
Specifically, we minimize the VQ reconstruction loss $\mathcal{L}_\text{VQ}$ and frame quality loss $\mathcal{L}_G$ along with our temporal ordering loss $\mathcal{L}_\text{T}$ to optimize T-VQ. 
At the same time, we also update the discriminator $D$ via the standard adversarial training by maximizing the discrimination loss $\mathcal{L}_D$. For video completion, the masking strategy $\mathcal{M}$ masks the video frames with the probability $p$ and then acquires guided frames from diverse time points. 
MMVG regards text and partial-missing video by $\text{Enc}^\text{M}$ for cross-modal fusion, and $\text{Dec}^\text{M}$ further predicts the visual tokens of frames autoregressively as the complete video. As a sequential generation process, we apply the teacher-forcing trick. Instead of our predicted $\hat{\text{z}}$, the ground-truth $\text{z}$ from the previous timestamp is fed to stabilize the training. Each video decoding loss $\mathcal{L}_t$ at timestamp $t$ is summed up as $\mathcal{L}_\text{M}$ to optimize MMVG. According to $\mathcal{L}_t$, we update $p$ for effective masking probability. The entire optimization object can be summarized as two phases:
\begin{equation}
\begin{split}
    \text{T-VQ:}& \min_{\text{Enc}^\text{Q}, \text{Dec}^\text{Q}, C, \text{FC}^\text{T}} \max_{D} \mathcal{L}_\text{VQ}+\mathcal{L}_G+\mathcal{L}_D+\mathcal{L}_\text{T}  \\
    \text{MMVG:}& \min_{\text{Enc}^\text{M}, \text{Dec}^\text{M}} \mathcal{L}_\text{M}
\end{split}
\end{equation}

\section{Experiments}
\subsection{Experimental Setup}
\noindent \textbf {Datasets.~}
As a new task, we consider diverse video scenes with natural instructions for \texttt{TVC}. \textbf{Kitchen}~\cite{damen2018kitchen} records 22K egocentric videos about kitchen activity, which have different lengths (4-16 frames) with narrations. \textbf{Flintstones}~\cite{gupta2018flintstones} contains 25K animation videos (15 frames) from \textit{The Flintstones}, where each video description includes the characters and their behavior.  \textbf{MUGEN}~\cite{hayes2022mugen} is built from agents playing CoinRun~\cite{cobbe2019coinrun}, which consists of 375K gaming videos (16 frames) with detailed text annotations. All videos in these three datasets are resized into 128x128. An overview is shown in Table~\ref{table:dataset} and Fig.~\ref{fig:tvc}(b). Since MMVG can unify various tasks, we also evaluate video generation/prediction on widely-used \textbf{UCF-101}~\cite{soomro2012ucf101} and \textbf{BAIR}~\cite{ebert2017bair}, video infilling on UCF-101 following RaMViD~\cite{hoppe2022ramvid}, and text-to-video generation on \textbf{MSRVTT}~\cite{xu2016msrvtt}.

\vspace{0.5ex}
\noindent \textbf{Evaluation Metrics.~}
We apply the following metrics to evaluate \texttt{TVC} results: 1) \textbf{FVD}~\cite{unterthiner2019fvd} computes the video features~\cite{carreira2017i3d} distance to the ground truth; 2) \textbf{RCS}~\cite{wu2021godiva} is the relative visual-text similarity to the instruction, compared to the ground-truth video. We fine-tune the CLIP model~\cite{radford2021clip} on each dataset and adapt it to the video scene for a more precise alignment. Apart from automatic metrics, we also conduct a human evaluation from aspects of video quality, instruction relevance, and ground-truth similarity. We sample 75 \texttt{TVP} results for each task and adopt MTurk\footnote{Amazon MTurk: \url{https://www.mturk.com}. Our studies have been cleared by the human subject committee as an IRB-exempt protocol.} to rank over baselines and our MMVG. To avoid the potential ranking bias, we hire 3 MTurkers for each sample of prediction, rewind, and infilling tasks.

\begin{table}[t]
\centering \tablestyle{2pt}{1.1}
    \begin{tabular}{ccccc}
        \toprule
        Dataset & Train / Val & \#Frame & \#Word & FPS \\
        \midrule
        \textbf{Kitchen}~\cite{damen2018kitchen} & 16,695 / 5,804 & 8.3 & 2.8 & 6 \\
        \textbf{Flintstones}~\cite{gupta2018flintstones} & 22,666 / 2,518 & 15 & 16.5 & 5 \\
        \textbf{MUGEN}~\cite{hayes2022mugen} & 362,239 / 12,848 & 16 & 20.6 & 5 \\
        \bottomrule
    \end{tabular}
    \vspace{-2ex}
    \caption{The statistics of our used datsets to evaluate \texttt{TVC}.}
    \label{table:dataset}
    \vspace{-3ex}
\end{table}

\vspace{0.5ex}
\noindent \textbf{Implementation Detail.~}
T-VQ contains ResBlocks~\cite{he2017resnet} as the visual auto-encoder ($\text{Enc}^\text{Q}$ and $\text{Dec}^\text{Q}$). The discriminator $D$ follows a similar architecture to $\text{Enc}^\text{Q}$. For the vector quantization, we use a patch size 16, where a 128x128 video frame transforms into 8x8 discrete visual tokens. There are 1024 vocabularies in the codebook $C$, and the hidden embedding size is 256. We adopt batch size 32 with a learning rate of 4.5e-6 to optimize T-VQ by Adam~\cite{kingma2015adam}. MMVG is built in an encoder-decoder manner, where $\text{Enc}^\text{M}$ is a transformer with 24 layers, 16 attention heads, and hidden embedding size 1024. $\text{Dec}^\text{M}$ adopts a similar setting with temporal window size 3 in VideoSwin. The initial sample rate $p$ of the masking strategy $\mathcal{M}$ is 0.9 with an adjusting rate $\alpha$ as 0.1. We optimize MMVG through the mixed precision~\cite{micikevicius2018fp16} with batch size 4 by Adam. The learning rate is also 4.5e-6. All experiments are implemented in PyTorch~\cite{paszke2017pytorch} and done on 8 NVIDIA A100 GPUs.

\begin{table}[t]
\centering \tablestyle{2pt}{1.1}
    \begin{tabular}{cccc}
        \toprule
        Method & \textbf{Kitchen} & \textbf{Flintstones} & \textbf{MUGEN} \\
        \midrule
        VideoDiff~\cite{ho2022video-diffusion} & 138.6 & 206.4 & 410.7 \\
        MCVD~\cite{voleti2022mcvd} & 119.9 & 183.8 & 400.2 \\
        TATS~\cite{ge2022tats} & \underline{115.5} & \underline{157.5} & \underline{386.4} \\
        MMVG & \textbf{109.1} & \textbf{127.6} & \textbf{368.6} \\
        \bottomrule
    \end{tabular}
    \vspace{-2ex}
    \caption{\textbf{FVD} results of \textbf{video generation} on our \texttt{TVC} datasets.}
    \label{table:vg-main}
    \vspace{-3ex}
\end{table}

\subsection{Main Results}
Table~\ref{table:main} shows the results of all text-guided prediction, rewind, and infilling for \texttt{TVC}. VideoMAE~\cite{tong2022video-mae} is built upon MAE~\cite{he2022mae} and reconstructs the missing video cubes, which performs \texttt{TVC} by masking all video frames except the first or the last (or both). TATS~\cite{ge2022tats}, the SOTA on video generation, also produces videos as discrete visual tokens. Since TATS can only consider the past through the autoregressive transformer, it requires specific training for each task. We have MMVG$^\text{U}$ as the unified model that can support all \texttt{TVC} tasks with a single training and MMVG$^\text{S}$ to further train for each prediction, rewind, and infilling. We treat TATS as our main baseline\footnote{Since we cannot receive feasible results after training diffusion methods for our \texttt{TVC}, we evaluate unconditional video generation in Sec.~\ref{sec:study}. We use this repo (\url{https://github.com/lucidrains/video-diffusion-pytorch}) as VideoDiff and the official repo as MCVD.} and study the importance of guided text.

\vspace{0.5ex}
\noindent \textbf{\texttt{TVP}rediction.~}
VideoMAE attempts to produce all frames simultaneously, which is difficult to maintain video temporal consistency, resulting in a high 328.9 FVD on Kitchen. TATS is inherently designed for prediction as it generates the frames one after one. However, our unified MMVG$^\text{U}$ performs better than TATS on all datasets (\textit{e.g.}, lower 105.6 and 124.8 FVD on Kitchen and Flintstones). These results support that learning from diverse time points will not hurt the prediction from the past. In contrast, our masking strategy can bring superior temporal coherence. MMVG$^\text{S}$ further improves itself through training prediction as completion from the head. However, there are too many possible outcomes from just the beginning, where the predicted results may not meet the expectation (\textit{e.g.}, a high 370 FVD on MUGEN). The instruction as guidance makes it related to the expected ground-truth result. We can let MUGEN run, jump, or collect coins as the textual descriptions to achieve more controllability, leading to a noticeable improvement (\textit{e.g.}, a lower 84.8 FVD by MMVG$^\text{U}$). The higher 70.2 RCS also shows that our MMVG can produce MUGEN videos that confirm with the instruction. Although the model may try to imagine the animation or the kitchen activity, the language hint can provide a clear goal to anticipate. Likewise, MMVG$^\text{U}$ with text surpasses TATS, even though it is not designed for prediction only. The specific trained MMVG$^\text{S}$ benefits the unified model for further improvement.

\vspace{0.5ex}
\noindent \textbf{\texttt{TVR}ewind.~}
Rewind from the last allows the model to imagine what happened along with a suitable opening. In addition, the objects may not display on the last frame (\textit{e.g.}, the spoons and forks for ``\textit{close drawer}''), which makes it more challenging to complete. Similar to prediction, VideoMAE cannot have feasible rewind results. Language is still essential to remind the past and establish an adequate beginning, where we can find a significant performance gap between with and without text (\textit{e.g.}, 90 \textit{vs.} 350 FVD on MUGEN). Our unified MMVG$^\text{U}$ achieves comparable results to TATS and even outperforms on Kitchen and Flintstones (\textit{e.g.}, higher 66.9 and 71.6 RCS). With the learning of completion from partial frames, autoregressive model can still accomplish video rewind without specific training. If following TATS design to train MMVG$^\text{U}$ for rewind, MMVG$^\text{S}$ gains more improvement and utterly surpasses it.

\vspace{0.5ex}
\noindent \textbf{\texttt{TVI}nfilling.~}
We consider the additional FILM~\cite{reda2022film} for infilling, which performs video interpolation with in-between motion. Despite synthesizing intermediate frames between the first and the last, the visual dynamics are changing too rapidly to handle, resulting in a higher FVD. With guidance from the head and tail, we find a noticeable improvement even without instruction (\textit{e.g.}, lower FVDs on Kitchen), which is helpful in temporal video modeling. To capture the expected missing event, we still require the language hint for more controllability. Our unified MMVG$^\text{U}$ achieves comparable performance to TATS again, which is specifically trained for the infilling task. It shows that completion from partial frames at different time points still helps, and MMVG$^\text{S}$ further outperforms on \texttt{TVI}nfilling.

\begin{figure}[t]
\centering
\begin{minipage}{.57\linewidth}
\centering \tablestyle{2pt}{1.1}
    \begin{tabular}[t]{ccc}
        \toprule
        ~ & \multicolumn{2}{c}{\textbf{UCF-101}} \\
        \cmidrule{2-3}
        Method & IS$\uparrow$ & FVD$\downarrow$ \\
        \midrule
        VideoGPT~\cite{yan2021video-gpt} & 24.7 & - \\
        DIGAN~\cite{yu2022di-gan} & 32.7 & 577 \\
        VideoDiff~\cite{ho2022video-diffusion} & 57.0 & - \\
        TATS~\cite{ge2022tats} & \underline{57.6} & \underline{420} \\
        MMVG & \textbf{58.3} & \textbf{395} \\
        \bottomrule
    \end{tabular}
    \vspace{-2ex}
    \captionof{table}{Results of \textbf{video generation} on UCF-101~\cite{soomro2012ucf101}.}
    \label{table:vg-ucf101}
\end{minipage}~~
\begin{minipage}{.42\linewidth}
\centering \tablestyle{2pt}{1.1}
    \begin{tabular}[t]{cc}
        \toprule
        ~ & \textbf{BAIR} \\
        \cmidrule{2-2}
        Method & FVD$\downarrow$ \\
        \midrule
        VideoGPT~\cite{yan2021video-gpt} & 103.3 \\
        MaskViT~\cite{gupta2022mask-vit} & 93.6 \\
        MCVD~\cite{voleti2022mcvd} & 89.5  \\
        TATS~\cite{ge2022tats} & \underline{88.6}  \\
        MMVG & \textbf{85.2} \\
        \bottomrule
    \end{tabular}
    \vspace{-2ex}
    \captionof{table}{Results of \textbf{video prediction} on BAIR~\cite{ebert2017bair}.}
    \label{table:vp-bair}
\end{minipage}
\vspace{-3ex}
\end{figure}

\subsection{Additional Study} \label{sec:study}
\noindent \textbf{Video Generation/Prediction.~}
We further evaluate the classic video generation and prediction tasks. Table~\ref{table:vg-main} shows FVD scores of unconditional video generation on our \texttt{TVC} datasets. Note that only videos but no texts are used in these experiments. Both VideoDiff~\cite{ho2022video-diffusion} and MCVD~\cite{voleti2022mcvd} are built upon denoising diffusion~\cite{ho2020diffusion}, where MCVD also considers different partial frames during training. The results first indicate that the vanilla token-based method is superior to the diffusion-based model  (TATS \textit{vs.} VideoDiff) for video generation. In addition, MMVG, with the masking strategy that learns the visual guidance from diverse time points, further boosts the performance (the lowest 127.6, 109.1, and 368.6 FVD on Flintstones, Kitchen, and MUGEN, respectively).

We also evaluate MMVG on UCF-101~\cite{soomro2012ucf101}, which is challenging to generate natural human videos. Table~\ref{table:vg-ucf101} supports that our MMVG can produce videos with higher visual similarity (a higher 58.3 IS) and temporal alignment (a lower 395 FVD) to the ground truth. For video prediction, we apply the widely-used BAIR~\cite{ebert2017bair} in Table~\ref{table:vp-bair}, where the model has to anticipate how a robot pushes objects from the given first frame. MMVG again surpasses TATS with the lowest 85.2 FVD. Although both generation and prediction are generating video frames chronologically, the ability to recover arbitrary missing frames for video completion empowers MMVG with a stronger temporal coherence, leading to better generative video quality.

\vspace{0.5ex}
\noindent \textbf{Video Infilling.~}
We follow RaMViD~\cite{hoppe2022ramvid} to evaluate video infilling on UCF-101. We consider various guidance settings $K$ in Table~\ref{table:vi-ucf101}. For example, $K$=+1 means given the first frame, and $K$=±2 should provide the first and last two frames. For prediction, MMVG outperforms RaMViD on all $K$, and the performance gap gets even larger when more guided frames are accessible (\textit{e.g.}, 33.4 on $K$=+1 and 65.9 on $K$=+5). A similar result can be found for infilling, where MMVG can make the lowest 120.3 FVD on $K$=±2. Despite having a similar masking strategy, it shows that generating frames one after one still brings superior results. MMVG allows autoregressive models to condition on visual hints from any time point, which produces more similar videos to the ground truth when infilling between the head and tail.

\begin{table}[t]
\centering \tablestyle{2pt}{1.1}
    \begin{tabular}{cccccc}
        \toprule
        ~ & \multicolumn{5}{c}{\textbf{UCF-101} (FVD$\downarrow$)} \\
        \cmidrule{2-6}
        Method & $K$=+1 & +2 & +5 & ±1 & ±2 \\
        \midrule
        RaMViD~\cite{hoppe2022ramvid} & 349.7 & 300.6 & 260.5 & 215.4 & 162.5 \\
        MMVG & \textbf{316.3} & \textbf{258.5} & \textbf{194.6} & \textbf{183.2} & \textbf{120.3} \\
        \bottomrule
    \end{tabular}
    \vspace{-2ex}
    \caption{Results of \textbf{video prediction} and \textbf{infilling} on UCF-101.}
    \label{table:vi-ucf101}
    \vspace{-3ex}
\end{table}

\vspace{0.5ex}
\noindent \textbf{Text-to-Video Generation.~}
Being a multimodal generative model, MMVG supports text-to-video generation. To compare with those large-scale methods, we pre-train MMVG using WebVid~\cite{bain2021frozen}, which contains 2.5M text-video pairs. We adopt the masking strategy to treat the pre-training as video completion. MMVG outperforms GODIVA~\cite{wu2021godiva} and NUWA~\cite{wu2022nuwa} without access the MSRVTT~\cite{xu2016msrvtt} data in Table~\ref{table:t2v-msrvtt}. Surprisingly, MMVG can generate videos that are more related to the texts (a higher 0.2644 CLIP-S~\cite{wu2021godiva}) than CogVideo~\cite{hong2022cog-video}, even though using twice less data. This result encourages the effectiveness of completion from partial frames. For a fair comparison without additional data, we directly train on MSRVTT. MMVG still outperforms TATS, which shows that text-to-video generation can be improved through learning from video completion as well.

\begin{table}[t]
\centering \tablestyle{2pt}{1.1}
    \begin{tabular}{cccccc}
        \toprule
        ~ & ~ & ~ & \multicolumn{2}{c}{\textbf{MSRVTT}} \\
        \cmidrule{4-5}
        Method & Pre-training & Zero-shot & FID$\downarrow$ & CLIP-S$\uparrow$ \\
        \midrule
        \textcolor{lightgray}{GODIVA}~\cite{wu2021godiva} & \textcolor{lightgray}{136M} & \textcolor{lightgray}{\ding{55}} & \textcolor{lightgray}{-} & \textcolor{lightgray}{0.2402} \\
        NUWA~\cite{wu2022nuwa} & 3.9M & \ding{55} & 47.7 & 0.2439 \\
        CogVideo~\cite{hong2022cog-video} & 5.4M & \ding{51} & 23.6 & 0.2631 \\
        \textcolor{lightgray}{Make-A-Video}~\cite{singer2022make-a-video} & \textcolor{lightgray}{20M} & \textcolor{lightgray}{\ding{51}} & \textcolor{lightgray}{13.2} & \textcolor{lightgray}{0.3049} \\
        MMVG & 2.5M & \ding{51} & \textbf{23.4} & \textbf{0.2644} \\
        \midrule
        TATS~\cite{ge2022tats} & - & \ding{55} & 63.2 & 0.2326 \\
        MMVG & - & \ding{55} & \textbf{60.6} & \textbf{0.2385} \\
        \bottomrule
    \end{tabular}
    \vspace{-2ex}
    \caption{Results of \textbf{text-to-video generation} on MSRVTT~\cite{xu2016msrvtt}. We gray out methods that use significantly more pre-training data.}
    \label{table:t2v-msrvtt}
    \vspace{-1ex}
\end{table}

\vspace{0.5ex}
\noindent \textbf{Human Evaluation.~}
We study the video quality (Q.), the relevance to the instructions (T.), and the similarity to the ground-truth video (GT) of the produced videos from the human aspect. The results in Table~\ref{table:human} are calculated as the mean ranking score (from 1 to 3, the higher is better) of each method for \texttt{TVP}rediction. MMVG without text even generates higher quality videos than TATS with text on Kitchen and MUGEN, where completion from partial frames benefits the temporal coherence of generative video modeling. While, the lowest ground-truth similarity illustrates that language guidance is crucial for controllability. With instruction, MMVG anticipates the future as the text (the highest T.) and generates videos that meet the ground truth (the highest GT), achieving the best overall performance.

\begin{table}[t]
\centering \tablestyle{2pt}{1.1}
    \begin{tabular}{cccccccccccccc}
        \toprule
        ~ & ~ & ~ &  \multicolumn{3}{c}{\textbf{Kitchen}} & ~ & \multicolumn{3}{c}{\textbf{Flintstones}} & ~ & \multicolumn{3}{c}{\textbf{MUGEN}} \\
        \cmidrule{4-6} \cmidrule{8-10} \cmidrule{12-14}
        Method & Text & ~ & Q. & T. & GT & ~ & Q. & T. & GT & ~ & Q. & T. & GT \\
        \midrule
        MMVG & \ding{55} & ~ & \underline{1.99} & 1.81 & 1.82 & ~ & 1.73 & 1.66 & 1.62 & ~ &  \textbf{2.03} & 1.56 & 1.55 \\
        TATS~\cite{ge2022tats} & \ding{51} & ~ & 1.97 & \underline{2.07} & \underline{2.03} & ~ & \underline{2.07} & \underline{2.12} & \underline{2.17} & ~ & 1.94 & \underline{2.11} & \underline{2.19} \\
        MMVG & \ding{51} & ~ & \textbf{2.04} & \textbf{2.12} & \textbf{2.15} & ~ & \textbf{2.20} & \textbf{2.22} & \textbf{2.21} & ~ & \underline{2.03} & \textbf{2.33} & \textbf{2.26} \\
        \bottomrule
    \end{tabular}
    \vspace{-2ex}
    \caption{Human evaluation for \texttt{TVP} with aspects of video quality (Q.), instruction relevance (T.), and ground-truth similarity (GT).}
    \label{table:human}
    \vspace{-3ex}
\end{table}

\begin{figure*}[t]
\centering
    \includegraphics[width=\linewidth]{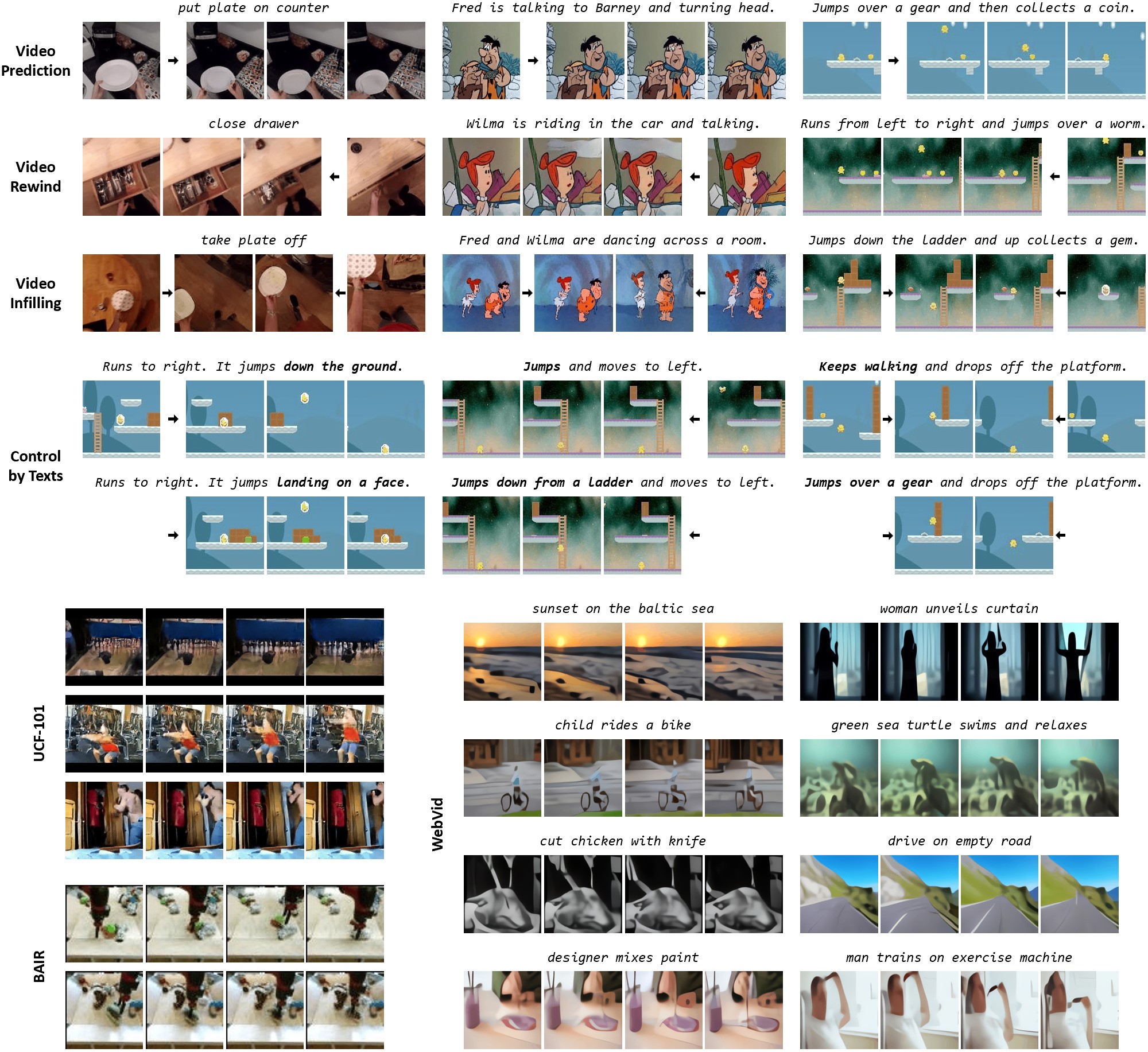}
    \vspace{-4ex}
    \caption{Qualitative examples of \texttt{TVC} on Kitchen~\cite{damen2018kitchen}, Flintstones~\cite{gupta2018flintstones}, and MUGEN~\cite{hayes2022mugen}. We also illustrate video generation on UCF-101~\cite{soomro2012ucf101}, video prediction on BAIR~\cite{ebert2017bair}, and text-to-video prediction on WebVid~\cite{bain2021frozen}.}
    \vspace{-3ex}
    \label{fig:qualitative}
\end{figure*}

\vspace{0.5ex}
\noindent \textbf{Qualitative Results.~}
Fig.~\ref{fig:qualitative} illustrates the keyframes of the generated examples of three datasets. Thanks to the learning of completion from partial frames at diverse time points, a single trained MMVG can support all \texttt{TVC} tasks. For prediction, MMVG makes Fred ``\textit{turn his head}'' or MUGEN ``\textit{jumps over the gear}'' from the guided text. MMVG further recovers the missing spoons and forks for ``\textit{close drawer}'' from the last frame in a more challenging rewind scenario. MMVG infills the missing event described in language such as ``\textit{stand up for dancing}'', ``\textit{walk across the kitchen}'', ``\textit{jump onto the stage}'' from the head and tail.

From the same visual guidance, MMVG can lead TVC results using different texts, achieving controllable video completion. For example, we can let MUGEN ``\textit{jump down the ground}'' or ``\textit{land on a face}'', starting from the same beginning. We can also control the behavior as ``\textit{keep walking}'' or ``\text{jump over a gear}'' to recover the missing middle event. Furthermore, MMVG also carries out unconditional video generation with smooth temporal coherence. We can use language to produce natural dynamics in diverse scenes, such as ``\textit{sunset on the sea}'', ``\textit{green sea turtle swins}'', or a close look of ``\textit{cut chicken}''. The presented videos indicate that our method not only unifies all \texttt{TVC} tasks but also performs the classical video generation/text-to-video well.

\section{Conclusion}
We introduce a novel task of text-guided video completion (\texttt{TVC}) that performs video completion from the first, last, or both frame(s) controlled by language. 
We present Multimodal Masked Video Generation (MMVG) with an effective masking strategy to learn the visual guidance from any time point. 
By varying the masking conditions, MMVG addresses all prediction, rewind, and infilling tasks within one model. 
Experiments on various video scenes show that our MMVG effectively addresses \texttt{TVC} as well as generative video modeling.
We believe \texttt{TVC} can help advance a new field toward vision-and-language research.

\clearpage

\appendix

\begin{figure}
\centering
    \includegraphics[width=\linewidth]{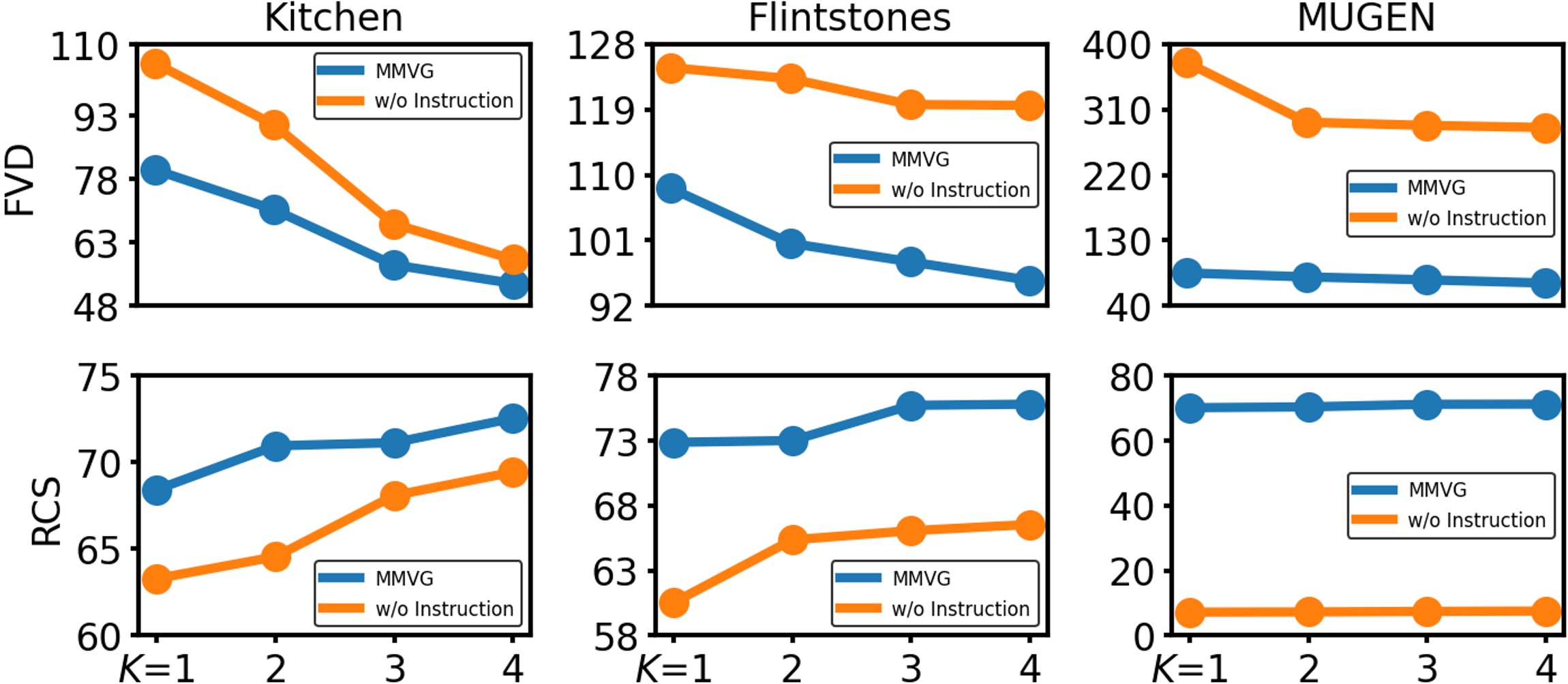}
    \vspace{-4ex}
    \captionof{figure}{Different \textbf{numbers of past frame} for \texttt{TVP}rediction\\(\textcolor{blue}{\textbf{MMVG}} \textit{vs.} \textcolor{orange}{\textbf{w/o Instruction}}).}
    \label{fig:different-k}
    \vspace{-1ex}
\end{figure}

\section{More Past Frames for \texttt{TVP}rediction}
As illustrated in Fig.~\ref{fig:different-k}, we explore the effect of different past frames ($K$) for \texttt{TVP}. With more past frames, we see a noticeable improvement for MMVG w/o Instruction (\textit{e.g.}, FVD decreases from 400 to 250 on MUGEN, and RCS increases from 63 to 68 on Kitchen). However, it is still far behind the one that has language guidance. Even only the first frame with the text outperforms using 4 past frames on Flintstones (\textit{e.g.}, a lower 110 FVD and a higher 73 RCS). Furthermore, MUGEN performs a series of actions and requires a longer temporal coherence; 4 past frames are insufficient to tell the expected outcome (\textit{e.g.}, the poor 7 RCS). The above results demonstrate the cruciality of instruction. On the other hand, for humans, supplying language is easier than drawing more video frames. Our \texttt{TVC} provides a practical setting that leads to effective video completion performance as well as human efficiency.

\begin{table}[t]
\centering \tablestyle{2pt}{1.1}
\resizebox{\linewidth}{!}{
    \begin{tabular}{ccccccccccc}
        \toprule
        ~ & ~ & ~ & \multicolumn{2}{c}{\textbf{Kitchen}} & ~ & \multicolumn{2}{c}{\textbf{Flintstones}} & ~ & \multicolumn{2}{c}{\textbf{MUGEN}} \\
        \cmidrule{4-5} \cmidrule{7-8} \cmidrule{10-11}
        Method & T-VQ & VidSwin & FVD$\downarrow$ & RCS$\uparrow$ & ~ & FVD$\downarrow$ & RCS$\uparrow$ & ~ & FVD$\downarrow$ & RCS$\uparrow$ \\ 
        \midrule
        TATS~\cite{ge2022tats} & - & - & 87.2 & 66.3 & ~ & 115.9 & 70.6 & ~ & 90.1 & 67.9 \\
        MMVG & \ding{55} & \ding{55} & \textbf{81.5} & \textbf{68.1} & ~ & \textbf{110.1} & \textbf{72.4} & ~ & \textbf{86.3} & \textbf{69.6} \\
        \hdashline
        \multirow{3}{*}{MMVG} & \ding{51} & \ding{55} & \underline{80.7} & \underline{68.3} & ~ & \underline{108.4} & \underline{72.6} & ~ & \underline{85.7} & \underline{70.0} \\
        ~ & \ding{55} & \ding{51} & 80.8 & 68.0 & ~ & 108.6 & 72.3 & ~ & 86.0 & 69.9 \\
        ~ & \ding{51} & \ding{51} & \textbf{80.2} & \textbf{68.4} & ~ & \textbf{108.2} & \textbf{72.9} & ~ & \textbf{84.8} & \textbf{70.2} \\
        \bottomrule
    \end{tabular}
}
    \vspace{-2ex}
    \caption{\textbf{Ablation study} of MMVG with temporal-aware VQGAN (T-VQ) and VideoSwin decoder (VidSwin) for \texttt{TVP}rediction.}
    \label{table:ablation}
    \vspace{-3ex}
\end{table}

\section{Ablation Study}
We conduct an ablation study to investigate each component effect in MMVG, including temporal-aware VQGAN (T-VQ) and VideoSwin decoder (VidSwin). T-VQ makes the reconstructed video from discrete tokens more smooth, and VidSwin considers latent temporal during the video decoding. If without T-VQ and VidSwin, MMVG will share a similar model architecture to TATS~\cite{ge2022tats} but contain the proposed masking strategy that learns video completion from arbitrary frames. In Table~\ref{table:ablation}, the performance gain mainly comes from the masking strategy (\textit{e.g.}, a lower 81.5 FVD on Kitchen and a higher 69.6 RCS on MUGEN), which validates the core idea of the mask-then-recover learning. Both T-VQ and VidSwin benefit the temporal coherence of video modeling, leading to an FVD decrease with a slight increase in RCS. In addition, combining all of them can bring a comprehensive improvement to MMVG.

\section{Detailed Analysis}
All experiments are conducted on the \textbf{Kitchen} dataset for \textbf{\texttt{TVP}rediction}. We have a detailed comparison among VQGAN~\cite{esser2021taming}, TA-VQ~\cite{ge2022tats}, and T-VQ. As shown below, our T-VQ outperforms the others on both frame reconstruction (over real video) and further \texttt{TVP}.
\begin{table}[H]
\centering \tablestyle{5.5pt}{1.0} \scriptsize
    \vspace{-2ex}
    \begin{tabular}{cccccc}
        \toprule
        ~ & \multicolumn{2}{c}{\textbf{Reconstruction}} & ~ & \multicolumn{2}{c}{\textbf{\texttt{TVP}rediction}} \\
        \cmidrule{2-3} \cmidrule{5-6}
        VQ Model & MSE$\downarrow$ & FVD$\downarrow$ & ~ & FVD$\downarrow$ & RCS$\uparrow$ \\
        \midrule
        VQGAN~\cite{esser2021taming} & 0.01582 & 36.72 & ~ & 82.3 & 66.4 \\
        TA-VQ~\cite{ge2022tats} & 0.00926 & 20.05 & ~ & 80.8 & 68.0 \\
        T-VQ (ours) & \textbf{0.00868} & \textbf{14.83} & ~ & \textbf{80.2} & \textbf{68.4} \\
        \bottomrule
    \end{tabular}
    \vspace{-3ex}
\end{table}

The Dec$^\text{Q}$ predicts the next frame based on the previous ones in an autoregressive manner. This enables a smooth transition and models the temporal dependencies, which is crucial for generative video modeling~\cite{ge2022tats,hong2022cog-video,yan2021video-gpt}. We also consider using parallel decoding, but it performs far below the autoregressive way.
\begin{table}[H]
\centering \tablestyle{5.5pt}{1.0} \scriptsize
    \vspace{-2ex}
    \begin{tabular}{ccc}
        \toprule
        Decoding & FVD$\downarrow$ & RCS$\uparrow$ \\
        \midrule
        Parallel & 113.5 & 64.6 \\
        Autoregressive & \textbf{80.2} & \textbf{68.4} \\
        \bottomrule
    \end{tabular}
    \vspace{-3ex}
\end{table}

For training, we have a high masking ratio $p$, with no more than 4 frames. Each frame is 64 tokens, so the length is no longer than 64*4+5([\texttt{SPAN}])+77(text)=340, which is efficient for current sequential modeling. During inference, there could be only two frames (head and tail) for the infilling task. We also compare the entire generation process (enc+dec+VQ) to VideoDiff~\cite{ho2022video-diffusion}. MMVG shows a better time/GPU efficiency for single and parallel inference.
\begin{table}[H]
\centering \tablestyle{5.5pt}{1.0} \scriptsize
    \vspace{-2ex}
    \begin{tabular}{cccccc}
        \toprule
        ~ & \multicolumn{2}{c}{\textbf{Time (sec)}} & ~ & \multicolumn{2}{c}{\textbf{GPU (MB)}} \\
        \cmidrule{2-3} \cmidrule{5-6}
        Model & BS=1 & 4 & ~ & BS=1 & 4 \\
        \midrule
        VideoDiff~\cite{ho2022video-diffusion} & 20.6 & 45.0 & ~ & 22419 & 37400 \\
        MMVG & \textbf{16.9} & \textbf{20.1} & ~ & \textbf{19814} & \textbf{32018} \\
        \bottomrule
    \end{tabular}
    \vspace{-3ex}
\end{table}

We also try applying the same cube embedding and optimize it via MSE loss. The result indicates that MMVG still surpasses VideoMAE~\cite{tong2022video-mae}. Moreover, the cube embedding cannot present clear and detailed pixels, which is unsuitable for video generation.
\begin{figure}[H]
\vspace{-2ex}
\centering
\begin{minipage}{.66\linewidth}
    \centering \tablestyle{5.5pt}{1.0} \scriptsize
    \begin{tabular}{cccc}
        \toprule
        Model & Output & FVD$\downarrow$ & RCS$\uparrow$ \\
        \midrule
        VideoMAE~\cite{tong2022video-mae} & Cube & 328.9 & 47.6 \\
        MMVG$^\text{U}$ & Cube & \textbf{272.6} & \textbf{50.7}  \\
        \midrule
        MMVG$^\text{U}$ & VQ & \textbf{105.6} & \textbf{63.3} \\
        \bottomrule
    \end{tabular}
\end{minipage}~~
\begin{minipage}{.2\linewidth}
    \centering
    \includegraphics[width=\linewidth]{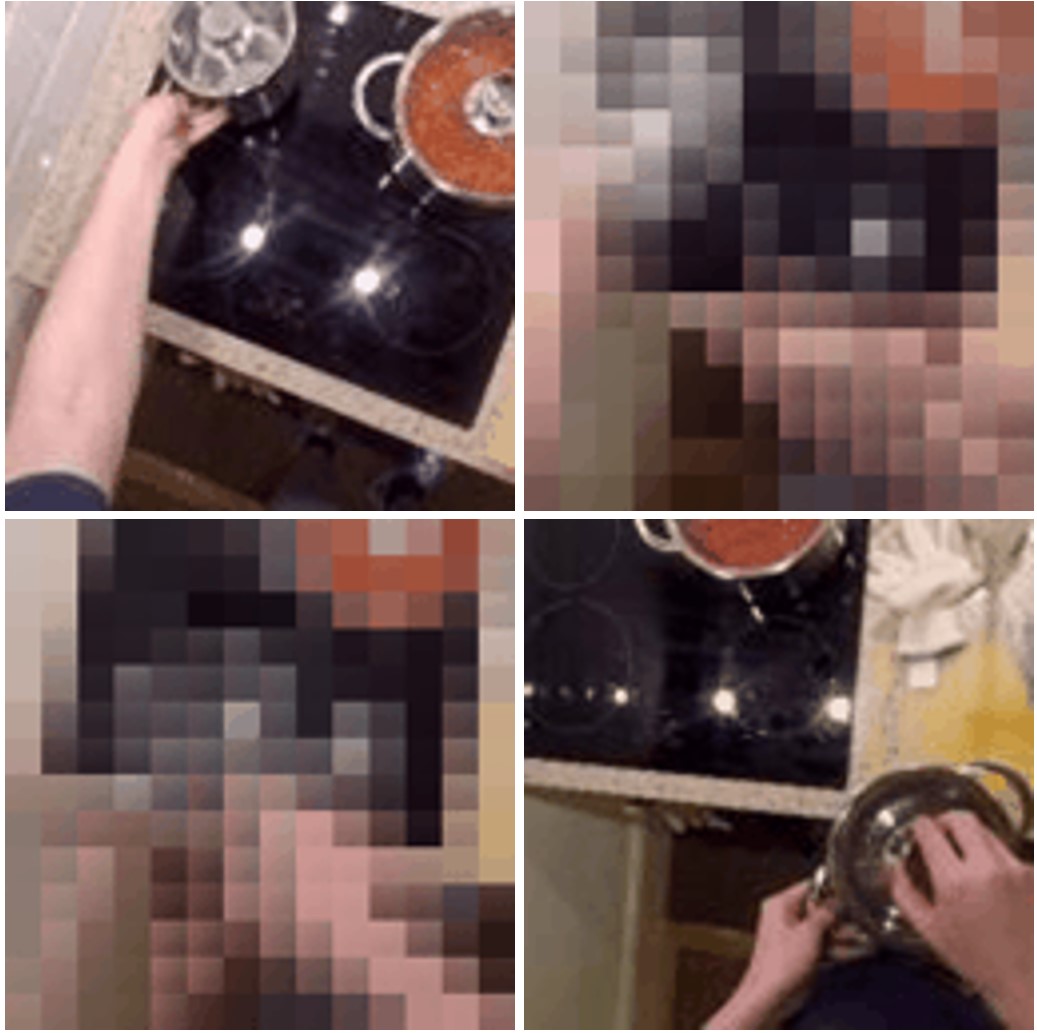}
\end{minipage}
\vspace{-2ex}
\end{figure}

For a fair comparison to TATS~\cite{ge2022tats}, we adopt the same 1024 codebook size and a 24-layer transformer. We consider the common setting,  8192 codes (as DALL-E) or 12 layers (as BERT-base). More VQ codes do not affect the quality. In contrast, to imitate the diverse activity motions, MMVG requires a larger model capability. Without teacher-forcing, the model cannot be trained effectively since it associates incorrect inputs with the corresponding outputs.
\begin{table}[H]
\centering \tablestyle{5.5pt}{1.0} \scriptsize
    \vspace{-2ex}
    \begin{tabular}{cccc}
        \toprule
        Codebook & \#Layer & FVD$\downarrow$ & RCS$\uparrow$ \\
        \midrule
        1024 & 12 & 99.3 & 65.2 \\
        1024 & 24 & \textbf{80.2} & \textbf{68.4} \\
        \midrule
        8192 & 24 & 79.9 & 68.3 \\
        \bottomrule
    \end{tabular}
    \vspace{-3ex}
\end{table}

For \texttt{TVC}, it is challenging to generate new objects that are not presented in the visual cues or the instruction. As the failure cases, though MMVG can make the motion of ``\textit{open the fridge}'', the items inside the fridge remain blurry. This highlights the need for human common sense to achieve realistic video generation.
\begin{figure}[H]
\centering
    \vspace{-2ex}
    \includegraphics[width=\linewidth]{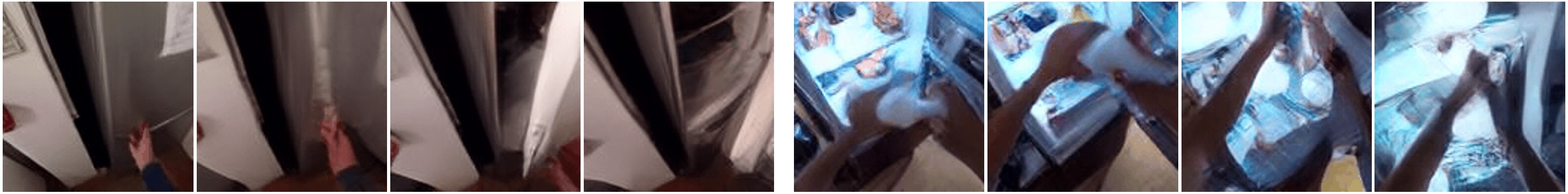}
    \vspace{-5ex}
\end{figure}

We also demonstrate video completion from intermediate, where $\nwarrow$ means the provided frames.
\begin{figure}[H]
\centering
    \vspace{-2ex}
    \includegraphics[width=\linewidth]{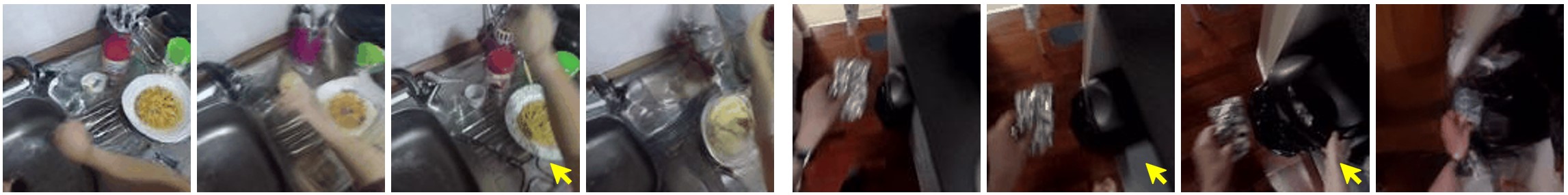}
    \vspace{-5ex}
\end{figure}

\begin{table}[t]
\centering \tablestyle{2pt}{1.1}
    \begin{tabular}{cccccccccc}
        \toprule
        ~ & ~ & \multicolumn{2}{c}{\textbf{Kitchen}} & ~ & \multicolumn{2}{c}{\textbf{Flintstones}} & ~ & \multicolumn{2}{c}{\textbf{MUGEN}} \\
        \cmidrule{3-4} \cmidrule{6-7} \cmidrule{9-10} 
        Method & FT. & R@1 & R@5 & ~ & R@1 & R@5 & ~ & R@1 & R@5 \\
        \midrule
        \multirow{3}{*}{CLIP~\cite{radford2021clip}} & \ding{55} & 2.0 & 7.4 & ~ & 23.0 & 45.0 & ~ & 0.2 & 1.6 \\
        ~ & Mean & \underline{11.4} & \underline{39.2} & ~ & \underline{73.2} & \underline{97.0} & ~ & \underline{11.4} & \underline{31.2} \\
        ~ & Temporal & \textbf{33.6} & \textbf{79.8} & ~ & \textbf{93.4} & \textbf{100} & ~ & \textbf{47.2} & \textbf{84.4} \\
        \bottomrule
    \end{tabular}
    \vspace{-2ex}
    \caption{Results of \textbf{instruction-to-video retrieval} by CLIP with different fine-tunings (FT.). We sample 1K pairs for this study.}
    \label{table:clip}
    \vspace{-3ex}
\end{table}

\section{Fine-tune CLIP as Evaluator}
The CLIP model~\cite{radford2021clip} has shown promising results by its strong text-visual alignment. GDOVIA~\cite{wu2021godiva} adopts CLIP and first proposes relative CLIP similarity (RCS) to evaluate text-guided visual generation. Since the video scene is in a specific domain and may differ from CLIP, we further fine-tune CLIP on each \texttt{TVC} dataset for a more precise alignment as our evaluator. We consider two fine-tuned settings: \textit{Mean} and \textit{Temporal}. \textit{Mean} applies a mean pooling layer over the visual features of all frames as the video features. On the other hand, \textit{Temporal} incorporates LSTM~\cite{hochreiter1997lstm} to acquire the temporal video features over frame features. Table~\ref{table:clip} presents the instruction-to-retrieval results with different fine-tuned settings. A higher recall represents a better alignment between instructions and videos. If directly using CLIP for RCS, it results in poor performance and is insufficient for our evaluation. By considering the latent temporal within video features, \textit{Temporal} leads to an overall advance and brings reliable alignment. We then finalize our evaluator as CLIP with the \textit{Temporal} fine-tuning.

\begin{table}[t]
\centering \tablestyle{2pt}{1.1}
    \begin{tabular}{cccc}
        \toprule
        ~ & ~ & \multicolumn{2}{c}{\textbf{UCF-101}} \\
        \cmidrule{3-4}
        Method & Pre-training & IS$\uparrow$ & FVD$\downarrow$ \\
        \midrule
        CogVideo~\cite{hong2022cog-video} & 5.4M & 50.5 & 626 \\
        \textcolor{lightgray}{Make-A-Video}~\cite{singer2022make-a-video} & \textcolor{lightgray}{20M} & \textcolor{lightgray}{82.6} & \textcolor{lightgray}{81} \\
        TATS~\cite{ge2022tats} & \ding{55} & 71.6 & 341 \\
        MMVG & \ding{55} & \textbf{73.7} & \textbf{328} \\
        \bottomrule
    \end{tabular}
    \vspace{-2ex}
    \caption{Results of \textbf{text-to-video generation} on UCF-101. We follow CogVideo~\cite{hong2022cog-video} to treat class labels as the input text. We gray out methods that use significantly more pre-training data.}
    \label{table:ucf}
    \vspace{-1ex}
\end{table}

\begin{figure}[t]
\centering
    \includegraphics[width=\linewidth]{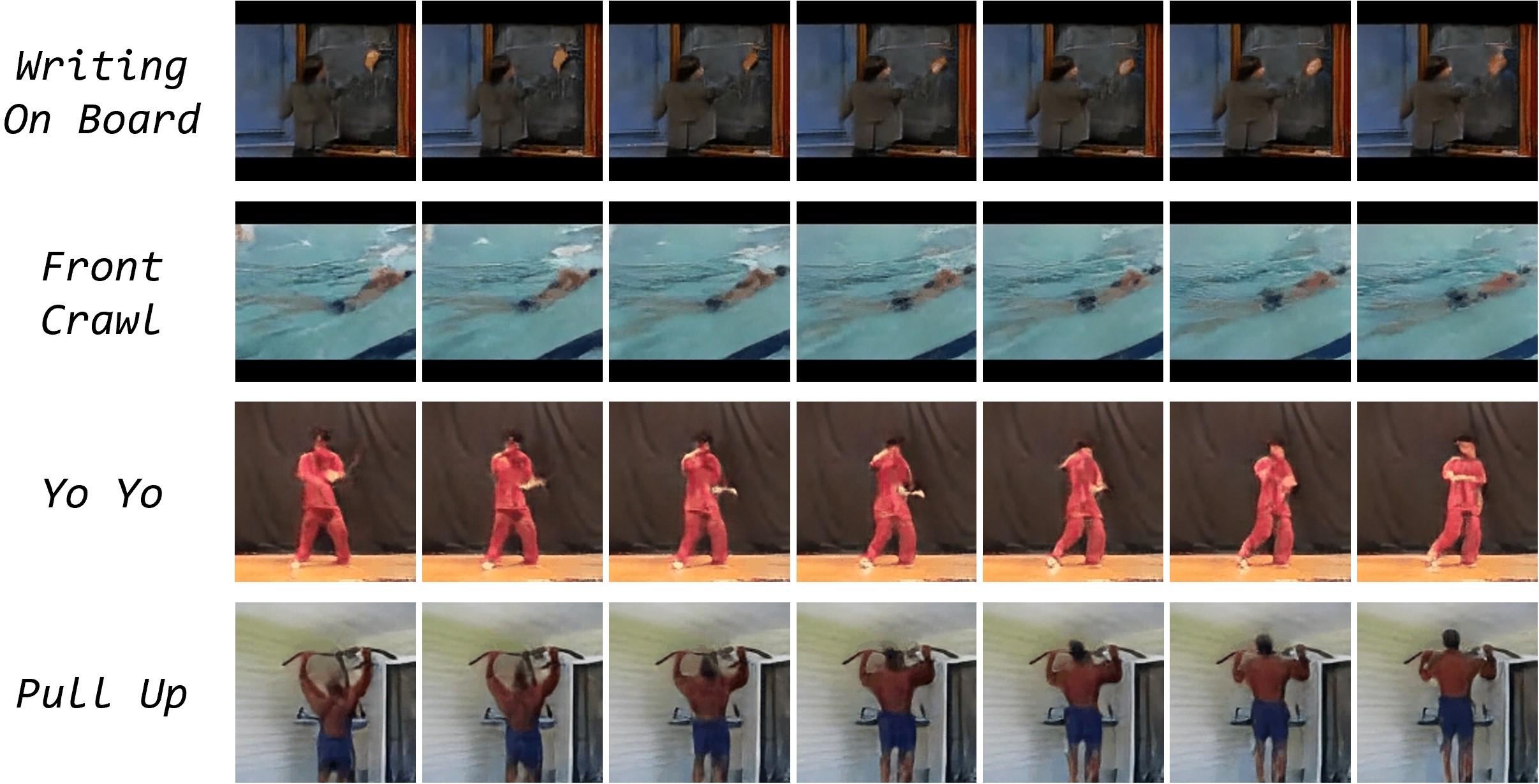}
    \vspace{-4ex}
    \caption{Qualitative examples of \textbf{text-to-video} on UCF-101.}
    \label{fig:ucf}
    \vspace{-3ex}
\end{figure}

\section{Text-to-Video Generation on UCF-101}
We follow CogVideo~\cite{hong2022cog-video} to treat class labels as the input text for text-to-video generation on UCF-101~\cite{soomro2012ucf101}. The results are shown in Table~\ref{table:ucf}. Our MMVG, without additional training data, can surpass large-scale pre-trained CogVideo. The higher 73.7 IS shows that the generated results are more diverse~\cite{salimans2016is}. And the lower 328 FVD also supports its better temporal coherence to ground-truth videos. When comparing MMVG to TATS, our masking strategy indicates the effectiveness that learning from completion can improve text-to-video. The qualitative examples are illustrated in Fig.~\ref{fig:ucf}.

\begin{figure}[t]
\centering
    \includegraphics[width=\linewidth]{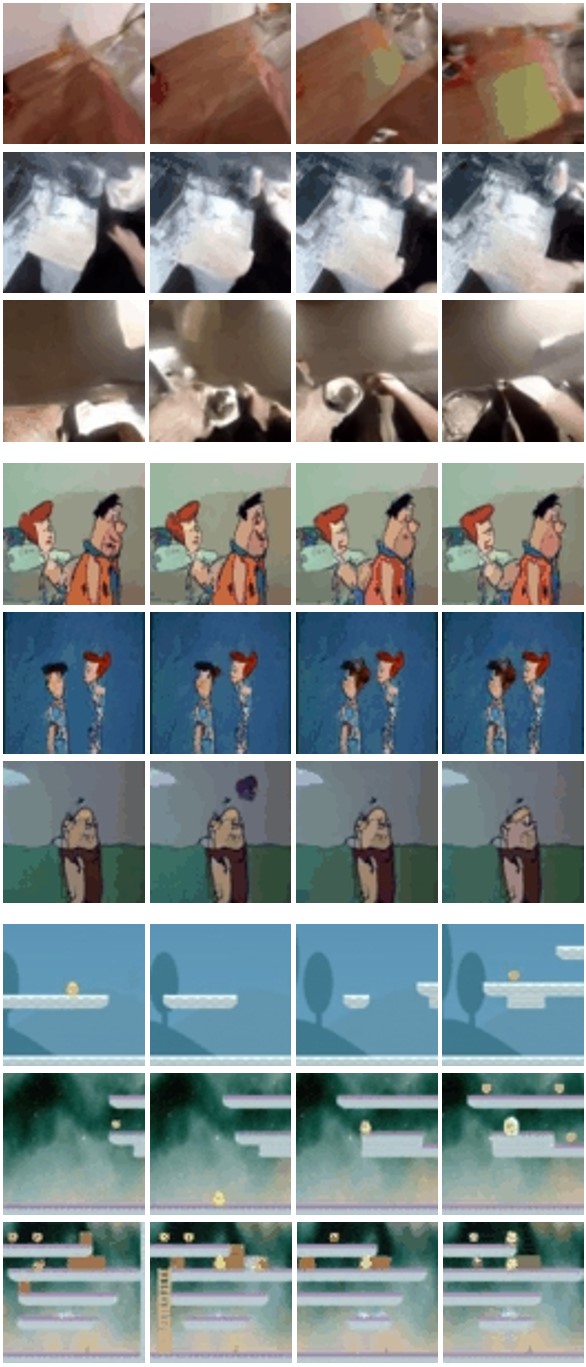}
    \vspace{-4ex}
    \caption{Qualitative examples of \textbf{unconditional video generation} on Kitchen, Flintstones, and MUGEN by VideoDiff~\cite{ho2022video-diffusion}.}
    \label{fig:diffusion}
    \vspace{-4ex}
\end{figure}

\section{Inferior Qualitative Results by VideoDiff}
We show that diffusion methods cannot generate as high-quality video as the used visual-token transformers (\textit{e.g.}, higher FVDs by VideoDiff~\cite{ho2022video-diffusion} and MCVD~\cite{voleti2022mcvd}). We further illustrate the qualitative examples by VideoDiff in Fig.~\ref{fig:diffusion}. As more challenging natural videos, we can see the blurring scenes on Kitchen. The motions are also unclear to tell what is actually doing. For Flintstones, it can produce characters but is difficult to present temporal dynamics, where the videos look almost static. Since it attempts to generate video frames from the 3D auto-encoder, VideoDiff cannot handle temporal coherence well. We still find obvious inconsistent results on MUGEN, even with the autoregressive video extension (\textit{e.g.}, the agent disappears with the platform being different lengths in the first case, or the ladder wrongly shows up in the third row.).

\begin{figure}
\centering
    \includegraphics[width=\linewidth]{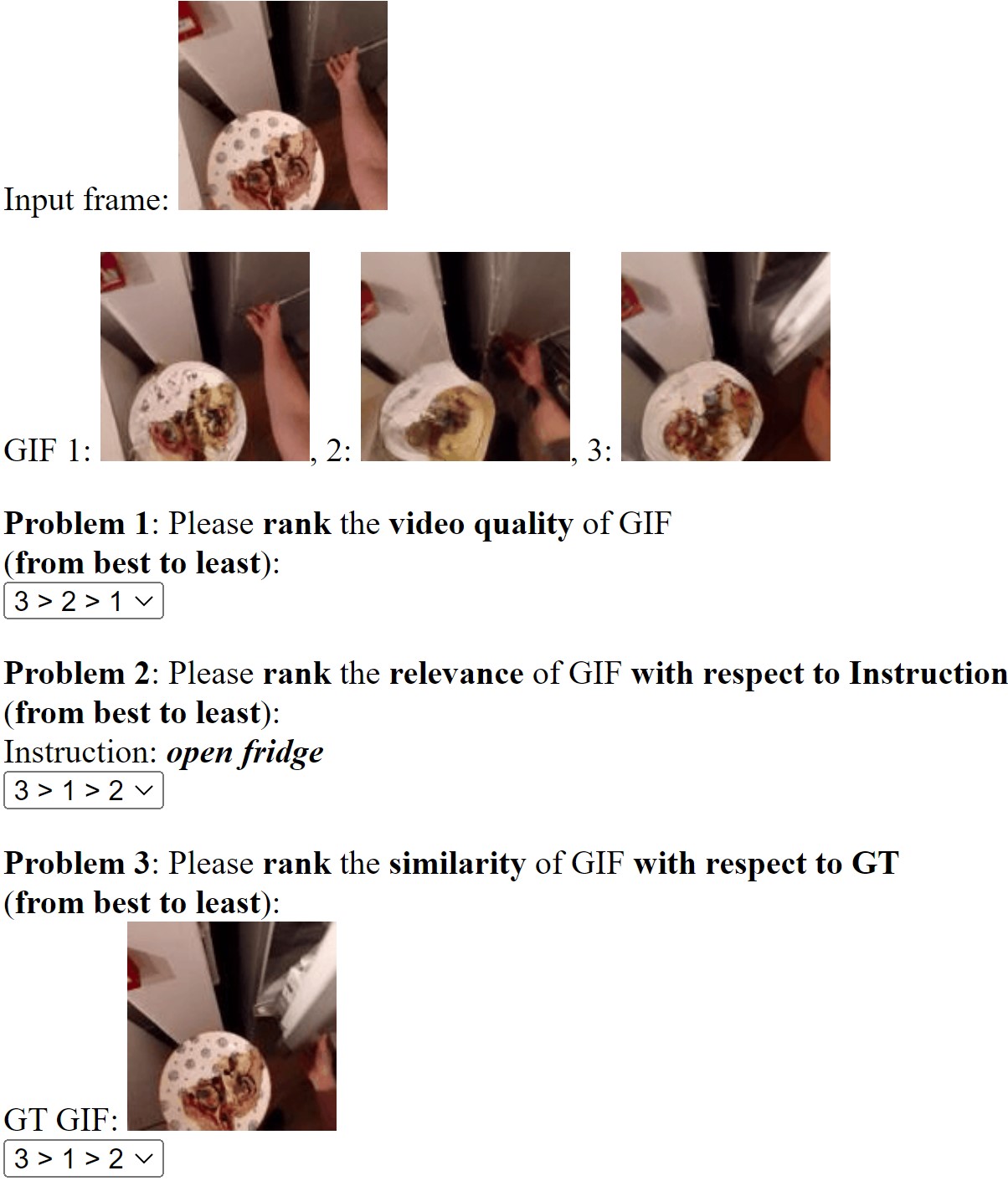}
    \vspace{-4ex}
    \captionof{figure}{Screenshot of the ranking tasks for human evaluation.}
    \label{fig:human}
    \vspace{-3ex}
\end{figure}

\section{Human Evaluation}
As illustrated in Fig.~\ref{fig:human}, we investigate the quality of generated results from the human aspect via Amazon Mechanical Turk. MTurkers rank the correlation of the \texttt{TVC} result concerning video quality, instruction relevance, or ground-truth similarity. Each MTurker rewards \$4.0 for a question group and takes a mean of 7 minutes.

\section{Social Impact and Ethics Discussion}
\texttt{TVC} brings out a general video completion that can generate a video from frames at arbitrary time points and control via natural language. Although our work benefits creative visual applications, there may be a ``\textit{fake as real}'' doubt for those produced videos. To mitigate this issue, we follow techniques in image forensics~\cite{wang2020ethic,frank2020ethic} and train a binary classifier~\cite{tran2018r2plus1d} to detect video authenticity. The accuracy on Kitchen, Flintstones, and MUGEN are all $>$99\%, which prevents them from counterfeiting. For guided instructions, hate speech detection~\cite{aluru2020ethic,caselli2021ethic} can be adopted to filter out potential malicious texts to avoid controversial results.

{\small
\bibliographystyle{ieee_fullname}
\bibliography{egbib}
}

\end{document}